\documentclass[]{article}
\usepackage[utf8]{inputenc}

\usepackage[ruled,linesnumbered]{algorithm2e}
\usepackage{hyperref}
\hypersetup{
	colorlinks=true,
	linkcolor=black,
	filecolor=black,      
	urlcolor=black,
	pdftitle={Supervised Dimensionality Reduction and Classification with Convolutional Autoencoders}
}
\usepackage[
backend=biber,
style=numeric,
sorting=ynt
]{biblatex}
\usepackage{graphicx}
\usepackage{subfig}
\usepackage{booktabs}
\newcommand{\ra}[1]{\renewcommand{\arraystretch}{#1}}

\title{Supervised Dimensionality Reduction and Image Classification Utilizing Convolutional Autoencoders}
\author{Ioannis A. Nellas\textsuperscript{1}, Sotiris K. Tasoulis\textsuperscript{1}, \\Vassilis P. Plagianakos\textsuperscript{1} and Spiros V. Georgakopoulos\textsuperscript{2}}
\date{\textsuperscript{1}\textit{Department of Computer Science and Biomedical Informatics\\
University of Thessaly, Greece}\\ \vspace{3ex} \textsuperscript{2}\textit{Department of Mathematics\\ University of Thessaly, Greece}\\ \vspace{3ex} \textit{ \{inellas, stas, vpp, spirosgeorg\}@uth.gr}\\ \vspace{3ex} August, 2022
}

\begin{document}

\maketitle

\begin{abstract}
    % The task of Image Classification is one of the elemental problems of computer vision and it is the basis of other well known problems of this field.
    The joint optimization of the reconstruction and classification error is a hard non convex problem, especially when a non linear mapping is utilized. In order to overcome this obstacle, a novel optimization strategy is proposed,
    in which a Convolutional Autoencoder for dimensionality reduction and a classifier composed by a Fully Connected Network, are combined to simultaneously produce supervised dimensionality reduction and predictions. 
    % composed by a Fully Connected Network, for the purpose of classifying the latent representation of the input is employed and thus achieving supervised dimensionality reduction and classification. 
    It turned out that this methodology can also be greatly beneficial in enforcing explainability of deep learning architectures. Additionally, the resulting Latent Space, optimized for the classification task, can be utilized to improve traditional, interpretable classification algorithms.
    % =======
    % Moreover, a thorough study of the Latent Space and the Classification behaviour of the proposed methodology is conducted.
    The experimental results, showed that the proposed methodology achieved competitive results against the state of the art deep learning methods, while being much more efficient in terms of parameter count. 
    % We additionally managed to improve traditional classification algorithms, when applied onto the latent representations of the proposed methodology.
    % Moreover, it was observed that the consideration of the label information onto the construction of the Latent Space, increased the separability of the data.
    Finally, it was empirically justified that the proposed methodology introduces advanced explainability regarding, not only the data structure through the produced latent space, but also about the classification behaviour. 
\end{abstract}

\section{Introduction}
    
    Image Classification, refers to the task of categorizing images into one of several predefined classes. This task is one of the fundamental problems in computer vision and it is the basis of other computer vision tasks such as segmentation and detection~\cite{rawat2017deep}. Traditionally, the procedure of image classification included two steps. First handcrafted image features were extracted via feature descriptors, such as SIFT~\cite{lowe2004distinctive} and SURF~\cite{bay2008speeded}, and then these features were used as input to a trainable classifier. The main limitation of this approach was the fact that the accuracy was heavily depended on the design of the feature extraction step, which was a time consuming and labor intensive task~\cite{rawat2017deep}. Thankfully, Deep Learning models and especially Convolutional Neural Networks (CNNs) have been shown to overcome this limitation and become the state of the art for image recognition, classification, and detection tasks ~\cite{heras2020supervised,rawat2017deep, mallat2016understanding}.
    
    In other cases, the images are directly treated as high-dimensional vectors where each variable correspond to an image pixel. Unfortunately, the resulting ultra  high-dimensional data, have some non intuitive characteristics. As presented in~\cite{aggarwal2001surprising}, the ratio of distances of a data point to its nearest and furthest neighbors tends to 1 as dimensionality grows, making not only the calculation of distances extremely computationally expensive but also affecting negatively the performance of classification methods. The emerged Dimensionality Reduction methods have been proven to be very effective in retaining the structure of data, making them a useful tool.
    % , even for classification purposes.

    Dimensionality Reduction is a widely used preprocessing step that facilitates classification, visualization and the storage of high-dimensional data~\cite{hinton2006reducing}. Especially for classification, it is utilised to increase the learning speed of the classifier, improve its performance and mitigate the effect of overfitting on small datasets through the noise reduction property of dimensionality reduction methods~\cite{wang2014role}. The majority of Supervised Dimensionality Reduction techniques, exploit the data and label pairs contained in the training dataset in order to learn the best dimensionality reduction mapping and then use those mappings as input to a standard classification algorithm. These methodologies, are the most common ones and they encourage the dimensionality reduction mapping to separate the inputs or manifolds that have different labels from each other, which can sometimes be effective~\cite{wang2014role}.
	
	Artificial Neural Networks have been widely used for dimensionality reduction as well. Autoencoder Networks, which are a nonlinear generalization of PCA~\cite{hinton2006reducing} have shown widespread success in producing powerful feature representations ~\cite{duan2019improving}. This type of network, allowed the process of feature extraction to be learnable and automatic, providing a flexible and scalable solution to the problem of dimensionality reduction and feature extraction~\cite{duan2019improving}. Most importantly, Autoencoders on the contrary to traditional dimensionality reduction methodologies, allow the utilization of deep learning architectures, such as convolutional networks, taking image local structure into consideration during feature extraction ~\cite{guo2017deep}.
	
% 	One method to improve the performance of classification is, instead of performing dimensionality reduction and then classification independently, best performance would be obtained by optimizing classification loss jointly with reconstruction error.
	
	In this work, we propose a methodology for supervised dimensionality reduction and classification based on a neural network architecture that optimizes classification loss jointly with reconstruction error, aiming to improve both classification performance and model explainability.
% ====
	The most well established example of this approach in statistical machine learning is the Linear Discriminant Analysis (LDA), which finds the best linear mapping, regarding between-class scatter against within-class scatter, that can also be used for classification tasks.
% 	An example of this type of approach, which finds the best linear mapping, regarding between-class scatter against within-class scatter, and then can be used for classification tasks, is Linear Discriminant Analysis (LDA).
	However, according to~\cite{wang2014role}, LDA solves a difficult non-convex problem, especially for a non-linear dimensionality reduction mapping. To overcome this obstacle, we propose a novel optimization strategy  which exploits a Convolutional Autoencoder for dimensionality reduction and a neural network classifier, entitled Convolutional Supervised Autoencoder (CSAE). Motivated by LDA, through this architecture we also focus on explainability, providing visualizations of the generated Latent Space. Furthermore, we show that the aforementioned Latent Space can greatly enhance the classification performance of traditional algorithms.  
% 	
% 	In this paper, in order to overcome this obstacle, a novel optimization strategy is proposed, which exploits a Convolutional Autoencoder for dimensionality reduction and a classifier, which is composed by a neural network, for the classification of the low dimensional representation of the inputs. This methodology is entitled Convolutional Supervised Autoencoder (CSAE). The latter, is also capable of providing explainability about its behaviour. Additionally, the Latent Space created by the proposed methodology, is optimized for classification and its powerful representations can be used as input to traditional classification algorithms in order to improve their performance.
	
% 	Furthermore, a study concerning the effect of the consideration of the label information onto the construction of the Latent Space of CSAE, is conducted. Finally, the Latent Space of CSAE and the classification behavior of the proposed methodology, are extensively studied, for the purpose of acquiring a deeper understanding of the way that the proposed methodology works.
	
% 	Finally, the effect of the consideration of label information during training onto the construction of the latent space of CSAE is studied.
	
	Hereby, the major contributions of this study are summarized:
	\begin{itemize}
	    \item A novel approach for supervised non linear dimensionality reduction and classification of image data, focusing on explainability through the generated Latent Space.
	    \item The utilization of the optimized for classification low dimensional representations from traditional classification algorithms to improve their performance.
	   %\item A study regarding the effect of the consideration of the label information onto the construction of the Latent Space of CSAE.
	    \item An extensive study of the resulting classification boundaries and their properties, through the resulting low dimensional representations. 
	   % \item A thorough study of the classification behavior of CSAE.
	    \item Provide extensive experimental analysis on real world benchmark and biomedical image datasets that justify the paper’s assumptions.
	   % \item Provide Visualizations of the Latent Space created by the proposed methodology.
	\end{itemize}

\section{Related Work}
    Images are characterized by high dimensionality, even when their size is relatively small, presenting a common case of the widely known curse of dimensionality. As described in ~\cite{aggarwal2001surprising} the ratio of distances of a data point to its nearest and furthest neighbors tends to 1 as dimensionality grows. This behaviour, affects negatively the performance of machine learning methods. A solution to this problem, came from the dimensionality reduction methods which have been proven to be effective on retaining the data structure, being a fruitful tool for the classification of high dimensional data~\cite{wang2014role}.
        
    The goal of dimensionality reduction is to retain as much of the significant structure of the high-dimensional data as possible in the low-dimensional representation. Principal Component Analysis (PCA), \cite{pearson1901liii}, projects the original data onto the directions of maximal variance in an unsupervised way. Linear Discriminant Analysis (LDA), as described in \cite{tharwat2017linear,wang2014generalized}, is a supervised dimensionality reduction method, which aims to find a linear subspace, which maximizes the between class to withing class variance ratio and thus guaranteeing maximum class separability.

    However, the manifold structure of real world data types, such as images, is complicated. As argued in~\cite{wang2014generalized}, the use of dimensionality reduction methods that utilize a simple parametric model, such as the Principal Component Analysis, or exploit fixed and defined data relations on the original high dimensional space, which may not be valid on the manifold, in order to learn the latter, e.g.\ ISOMAP~\cite{tenenbaum2000global}, are not sufficient to capture such complicated structures. Thankfully, Neural Networks and especially Autoencoders,  have shown widespread success in producing powerful feature representations~\cite{duan2019improving}, mitigating the previously presented limitation.
    
     % ========
    The Autoencoder Neural Networks~\cite{rumelhart1985learning}, is a non-linear generalization to Principal Component Analysis~\cite{hinton2006reducing}. This type of Neural Networks, have shown wide success as tools for non linear dimensionality reduction and feature extraction for clustering~\cite{duan2019improving,guo2017improved,xie2016unsupervised,nellas2021convolutional,makhzani2015adversarial, MRABAH2020206}, semi supervised learning~\cite{gogna2016semi,rasmus2015semi} and classification \cite{le2018supervised, nousi2020self, rolfe2013discriminative, gao2015single} tasks.
    % --
    % --
    In our interest, the Supervised Autoencoder (SAE), which is proposed in~\cite{le2018supervised}, is a neural network that jointly predicts the input and the classification result. Moreover, in the aforementioned study, a proof of the uniform stability of the SAE with one hidden layer (linear SAE), is performed, and thus a bound on the generalization error is provided. Finally, they empirically show that the addition of the reconstruction loss never harms performance when compared with the corresponding neural network.
    % --
    % --
    
    Another recent methodology that utilizes Autoecoders for classification is presented in~\cite{nousi2020self}. Therein, the Latent Space of the Autoencoder is exploited to perform classification while, a fine-tuning of the learned representation is performed in a self-supervised fashion, forcing the Autoencoder to learn better separated low dimensional representations. In an earlier study~\cite{rolfe2013discriminative}, the discriminative recurrent sparse auto-encoder model is proposed, which is composed of a recurrent encoder that has Rectified Linear Units and it is connected with two linear decoders which not only reconstruct the input but also predict the classification result. Simultaneously, the label information was embedded into the training of Autoencoder by enlarging the error function to include the classification error. Moreover, supervised deep autoencoders were used for Face Recognition~\cite{gao2015single}. Finally, in the study of~\cite{heras2020supervised}, extracted image features from pretrained Convolutional Neural Networks were exploited and provided as input to Linear Discriminant Analysis in order to perform Supervised Dimensionality Reduction and Classification.

    In this work, Supervised Dimensionality Reduction and Classification is studied similarly to~\cite{le2018supervised}. However, motivated by the aforementioned approaches more emphasis was laid on deep learning architectures for the Image Classification task.
      % Shmeiosi apo k. Tasouli me mple
%      \textcolor{blue}{\textbf{Note}: edo isos prepei na poume ti den kanoun kala kai ti diaforetiko kanoume emeis os pros ti methodologia pio ksekathara!. \textbf{Keimeno pou prostethike}:}
      In contrast to~\cite{le2018supervised}, a novel optimization strategy is proposed, while extensive visualizations of the generated Latent Space and its exploration are provided in order to deeply understand the behaviour of the proposed methodology in terms of performance and explainability  concerning its decision making, structure preservation and information capture behaviour. Additionally, the proposed methodology is characterized by lower complexity than the supervised Autoencoder presented in~\cite{nousi2020self}, while concurrently, as opposed to~\cite{rolfe2013discriminative}, pretraining is not required.
    In addition, we study the exploitation of the optimized for classification, Latent Space of the Convolutional Supervised Autoencoder in order to improve already existing classification algorithms. More importantly, the Latent Space, the classification behavior and the explainability of the proposed methodology, are extensively studied.
    
\section{Proposed Methodology}
    % Inspired by Linear Discriminant Analysis (LDA) and the methodology proposed in \cite{le2018supervised}, in this paper,
    In what follows, we present a novel optimization strategy for classification and reconstruction error. We exploit a Convolutional Autoencoder for dimensionality reduction that preserves local structure of data generating distribution, as presented in \cite{guo2017improved}, and a classifier, in the form of a Fully Connected Neural Network, in order to achieve the desired task. This methodology, provides a framework for supervised non-linear dimensionality reduction and classification in an end to end manner. The aforementioned methodology, is entitled Convolutional Supervised Autoencoder (CSAE). 
    Subsequently, 
    % the proposed methodology, creates a Latent Space of a much lower dimensionality, which is optimised for classification, because the latter is the main training objective. Thus, 
    we utilize the powerful latent representations of images that lie in the generated Latent Space, and use them as inputs to traditional classification algorithms, such as the k-Nearest Neighbors method, in order to improve their performance.
    
    \subsection{Convolutional Supervised Autoencoder}

    % Hypothesis: A non linear data transformation creates a space where the data are linearly separable.
    % Thus a linear classifier is able to solve the problem adequately in this space. (DONE)
    
     %Phrase: In order to examine non linear dimensionality reduction, the following hypothesis is formulated:{\em A non linear data transformation creates a space which the data are linearly separable.} (DONE)

    The primary tasks of this methodology are Supervised Dimensionality Reduction and Image Classification. The Convolutional Autoencoder, and thus the reconstruction error, is employed as an auxilary task in order to not only preserve the local structure of the data generating distribution, as presented in \cite{guo2017improved}, but also to act as a regularizer for the solution. This results in promoting stability and achieving better generalization \cite{le2018supervised}. In order to examine the non linear dimensionality reduction and classification capabilities of the proposed methodology, the following hypothesis is formulated and studied: {\em A non linear data transformation generates a space on-top of which the data are linearly separable.}
    
\begin{figure}[t!]
        \centering
        \includegraphics[width=\linewidth]{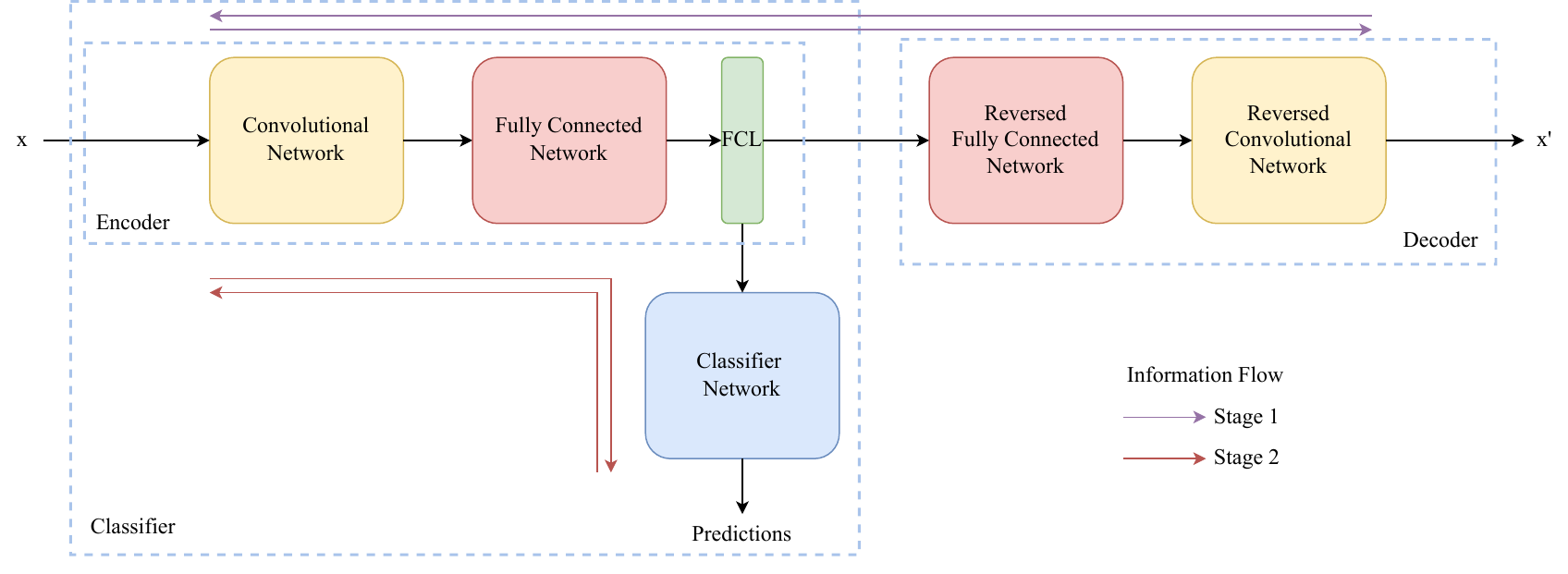}
        \caption{Schematic Representation of CSAE.}
        \label{fig:prop_methodlogy}
    \end{figure}

    A Graphical Representation of the proposed methodology is presented in Figure \ref{fig:prop_methodlogy} while, the proposed methodology for supervised dimensionality reduction and classification is presented in Algorithm \ref{alg:prop_meth}. In what follows, we describe one iteration, for a given batch of images and their labels:
    \begin{itemize}
        \item Initially, a forward pass of the batch of training images through the Convolutional Autoencoder is performed.
        \item The Loss Function of the Convolutional Autoencoder is evaluated, and its weights are updated through backpropagation.
        \item Then, a forwards pass of the batch of training images through the Classifier network is performed.
        \item Finally, the Loss Function of the Classifier is evaluated, using the image labels contained in the batch, and its weights are updated through backpropagation.
    \end{itemize}
    
    \begin{algorithm}[t!]
        \DontPrintSemicolon
        \caption{Image Classification with CSAE.} \label{alg:prop_meth}
        \KwData{$x_{train}$: The images in the train set,\\ $y_{train}$: The ground truth of images in the train set,\\ $x_{test}$: The images in the test set,\\ $epochs$: The number of epochs,\\ $W_{ae}$: The weights of the Convolutional Autoencoder,\\ $W_{cl}$: The weights of the Classifier Network.}
        \KwResult{$c_{out}$: the classification result of $x_{test}$.}
         \For{$epoch \leftarrow 1...epochs$}{
            Create $batches$ from $(x_{train},y_{train})$.\;
            Shuffle the $batches$.\;
            \For{$x^{batch}_{train}, y^{batch}_{train}$ $\in$ batches }{
            Forward pass of $x^{batch}_{train}$ through the Convolutional Autoencoder.\;
            Evaluate the Loss Function of the Convolutional Autoencoder.\;
            Update $W_{ae}$ by standard back propagation.\;
            Forward pass of $x^{batch}_{train}$ through the Classifier Network.\;
            Evaluate the Loss Function of the Classifier Network, using $y^{batch}_{train}$ as the ground truth of $x^{batch}_{train}$.\;
            Update $W_{ae}$ by standard back propagation.\;
            }
         }
         Detach the Classifier Network from CSAE.\;
         Pass the $x_{test}$ through the Classifier Network.\;
         Acquire $c_{out}$.\;
    \end{algorithm}
    
    % Meta apo thn suzhthsh me ton k. Plagianako prostethike protasi anaforika me to efficiency.
    This procedure is repeated until convergence or for a specified number of epochs. Subsequently, the Classifier Network is detached, to be used as a standalone classifier, reducing significantly the number of required parameters for the classification task. Finally, as Loss Functions of the Convolutional Autoencoder and the Classifier Networks, the Mean Squared Error (MSE) and the Categorical Crossentropy are utilized.

    \subsection{Improving Classification Methods with CSAE}
    
    % If the formulated hypothesis holds, then the exploit of a linear classifier onto that space should perform adequately, because as previously described the classes would be linearly separable. (DONE)
    
    One of the basic components of CSAE is the Convolutional Autoencoder. We consider the generated Latent Space \ref{alg:prop_meth} optimized for the classification task, since the minimization of the classification error is the main training objective.
    % 
    % As a consequence, label information was taken into consideration during the latent space construction.
    Additionally, this Latent Space is constrained by the reconstruction error of the Convolutional Autoencoder and thus, the local structure of the data generating distribution is preserved. Therefore, the feature space corruption phenomenon is mitigated, as described in \cite{guo2017improved}. Finally, it can be concluded that if the formulated hypothesis holds, then the exploit of a linear classifier onto the Latent Space of CSAE should perform adequately.
    % , since, as previously described, the classes would be linearly separable. 
    
    \begin{algorithm}[b!]
        \DontPrintSemicolon
        \caption{Improving Classification Methods with CSAE.} \label{alg:encoder_traditional}
        \KwData{$x_{train}$: The images in the train set,\\ $y_{train}$: The ground truth of images in the train set,\\ $x_{test}$: The images in the test set.}
        \KwResult{$c_{out}$: the classification result of $x_{test}$.}
         Train CSAE as described in Algorithm \ref{alg:prop_meth}. \;
         Acquire the Encoder Network from the trained CSAE. \;
         Pass $x_{train}$ and $x_{test}$ through the Encoder Network and acquire their low dimensional representations $z_{train}$ and $z_{test}$ respectively.\;
         Train a Traditional Classification Algorithm with $z_{train}$ and $y_{train}$.\;
         Acquire $c_{out}$ from the trained traditional classifier by providing $z_{test}$ as input.\;
    \end{algorithm}
    
\begin{figure}[t!]
        \centering
        \includegraphics[width=\linewidth]{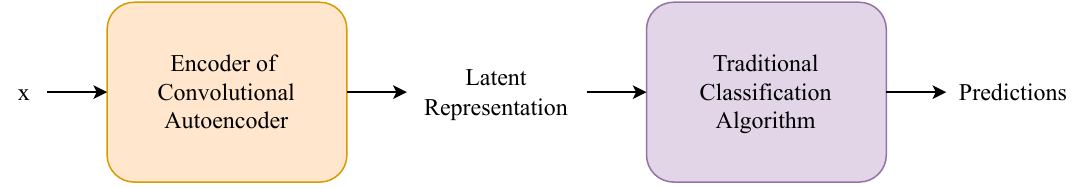}
        \caption{Schematic Representation of the methodology for improving Traditional Classification Algorithms using CSAE.}
        \label{fig:encoderTraditional}
    \end{figure}

    A Schematic Representation of the described methodology is presented in Figure \ref{fig:encoderTraditional} while the complete algorithmic procedure is presented in Algorithm \ref{alg:encoder_traditional}. In detail, CSAE is initially trained following the procedure described in Algorithm \ref{alg:prop_meth}. Then, the Encoder Network of the Convolutional Autoencoder is detached and a pass of the images contained in the train and test set through the Encoder Network is performed, in order to acquire their low dimensional representation, $z_{train}$ and $z_{test}$ respectively. Afterwards,  a traditional classifier is trained using $z_{train}$ and $y_{train}$. The classification result of the images contained in the test set $c_{out}$, is acquired by providing $z_{test}$ as input to the trained traditional classifier. 
    
    % ====================================
    % ====================================
    % Meta apo thn suzhthsh me ton k. Plagianako prostethike mia paragrafos anaforika me to efficiency tis methodou
    The advantages of this methodology are two fold. Initially, the only component of CSAE that is required for prediction is the Encoder Network, which means that the essential number of parameters are further reduced. Also, the second advantage is that, the original images can be deleted after the computation of their latent representations, and thus decreasing the memory requirements of the dataset, while concurrently increasing the execution time of the utilized traditional classification methodology. The images can be reconstructed by providing their latent representations as input to the Decoder Network. In conclusion, from the previously described advantages, it is realized that this methodology offers an efficient solution to the classification problem. 
    
\section{Experimental Analysis}
    This Section is devoted to the experimental evaluation of the proposed methodologies. For this purpose we employ two widely used benchmark datasets and two recent real world biomedical image datasets. In what follows, we provide a brief overview of the datasets, the preprocessing procedure and the evaluation metrics. In addition, we present the algorithms used for comparison and the experimental procedure. Finally, the experimental results are presented and interpreted through a thorough discussion.
    
    % In this section the experimental results of the aforementioned methodologies applied onto two benchmark and two biomedical image datasets are presented.
    % Specifically, a brief overview of the image datasets, the preprocessing procedure and the evaluation metrics is performed. In what follows, the algorithms used for comparison and the implementation of the experiments are described. Finally, the experimental results are presented and interpreted through a thorough  discussion.
    
\subsection{Datasets}
    Selecting widely used datasets for our experiments allow us to provide direct comparisons with recent methodologies found in the literature. For this purpose we utilized the MNIST and Fashion-MNIST dataset respectively. Nevertheless, we also utilize two recent biomedical image datasets to expose the true potential of the proposed methods both in terms of classification performance and their generalization capability. In detail, the four employed datasets are the following:
    % Experiment took place utilizing 2 benchmark image datasets,
    % The proposed methodologies were applied onto 2 benchmark image datasets, which they are widely used for the evaluation of classification methods, and two biomedical image datasets. The datasets that were used are the following:
    \begin{itemize}
        \item MNIST~\cite{lecun2010mnist}: is a dataset of 70,000 grayscale images of handwritten digits 0 to 9. Each image, contained in this set of data, has $28 \times 28$ pixels size.
		\item Fashion-MNIST~\cite{xiao2017fashion}: consists of 70,000 grayscale images, were each one is associated with a label from 10 classes. Each image has $28 \times 28$ pixels size.
		\item Brain Tumor Image Dataset~\cite{Cheng2017}: This dataset contains 3064 T1-weighted contrast-enhanced images from 233 patients with three kinds of brain tumor: meningioma (708 slices), glioma (1426 slices), and pituitary tumor (930 slices). This dataset is publicly available in Kaggle\footnote{see:\url{https://www.kaggle.com/denizkavi1/brain-tumor}}.
		\item SARS-COV-2 CT-Scan dataset~\cite{soares2020sars}: This dataset contains 2482 CT scans in total, where 1252 CT scans are positive for SARS-CoV-2 infection (COVID-19) and 1230 CT scans are from patients that are non-infected by SARS-CoV-2. This dataset was collected from real patients, which they are hospitalized in Sao Paulo, Brazil and it is publicly available in Kaggle\footnote{see:\url{https://www.kaggle.com/plameneduardo/sarscov2-ctscan-dataset?select=non-COVID}}. 
    \end{itemize}
    
\subsection{Data Preprocessing and Evaluation Metrics}

    % The images of the previously described datasets were preprocessed prior to the application of each methodology. More precisely, 
    The images contained in the Brain Tumor and SARS-COV-2 CT Scan datasets, were resized to $128 \times 128$ using the Nearest Neighbor Interpolation method. In addition, they were flattened and standardized for the application of traditional classification algorithms, and normalized to the $\left[0,1\right]$ range, for the rest methodologies. For the MNIST and Fashion MNIST datasets the provided train-test splits were used, while for the Brain Tumor Dataset and the SARS-COV-2 CT-Scans Datasets, the train and test splits were retrieved by random sampling an 80\% to 20\% ratio respectively. The validation set for each dataset was created by random sampling 10\% of samples from the training set. Finally, for the evaluation of the performance of the classification algorithms, two standard metrics were used: the Accuracy~\cite{grandini2020metrics} and the weighted by support F1-Score\footnote{as described in \url{https://scikit-learn.org/stable/modules/generated/sklearn.metrics.f1_score.html}}.

\subsection{Algorithms used for comparison}
    
    Aiming at the evaluation of the classification performance of CSAE, a wide variety of algorithms were used through an extensive comparison (namely, First set of comparisons). We initially compare against Linear Discriminant Analysis (denoted as LDA), in order to compare CSAE with the most well established methodology for Supervised Dimensionality Reduction and Classification.
    Subsequently, our aim is to illustrate the impact of each individual component of the proposed methodology to the classification result.
    Specifically, CSAE is compared with a CNN classifier (denoted as CNN classifier) of the same architecture, while also compared against the independent use of a Convolutional Autoencoder for dimensionality reduction and a Classifier (denoted as AE + Class. Net.), where both parts have similar architecture to the corresponding component of CSAE. Finally, CSAE is compared against several state of the art methodologies proposed in ~\cite{kabir2020spinalnet,nokland2019training,8451379,8683759, wang2020contrastive, Jaiswal2021Class}.

    To investigate the methodology that exploits the Latent Space of CSAE in order to improve the performance of traditional classification methodologies (Second set of comparisons), three traditional classification algorithms were used: k-Nearest Neighbors (denoted as k-NN), Support Vector Machines with Radial Basis Function Kernel (denoted as SVM) and the Naive Bayes Classifier (denoted as GNB). The comparison includes the execution of these methodologies to both the original flattened images,and the latent representations of the images created by CSAE, denoted as k-NN/SVM/GNB and CSAE L.S.+ k-NN/SVM/GNB respectively.

\subsection{Implementation}
    
    % ADDED: GNU General Public License v3.0
    The implementation of the whole experimental process was accomplished with the Python programming language and it is available under an open source licence through a GitHub page. The deep learning models, were implemented using the Keras~\cite{chollet2015keras} application programming interface (API), while for the traditional Machine Learning algorithms for classification and Evaluation Metrics, the implementations contained in Scikit-learn~\cite{scikit-learn} were utilized. The Convolutional Networks are constructed similarly to~\cite{guo2017deep}. More precisely, for the MNIST and Fashion MNIST datasets, the Convolutional Network of the Encoder, is composed of 2 convolutional layers, where the kernel maps were assigned to $3 \times 3$, while concurrently the number of filters were assigned to 32, 64 respectively. Also, for the Brain Tumor and SARS-COV-2 CT-Scan datasets, 4 convolutional layers were exploited, where the kernel maps were assigned to $5 \times 5$ for the first two convolutional layers and $3 \times 3$ for the last two, while simultaneously the number of filters were assigned to 32, 64, 128 and 256 respectively. The convolutional layers of the Convolutional Network, contained in the Decoder are identical to the encoder’s, but in reverse order. Additionally, the stride parameter for all the convolutional layers is set to two, because, as described in~\cite{guo2017deep}, this setting allows the convolutional network of the Encoder and its transpose counterpart contained in the Decoder to learn spatial subsampling and upsampling respectively and thus leading to higher capability of transformation. 
    
    The Fully Connected Network of the Encoder Network is composed of three fully connected layers. The first two were assigned to 128 units and the final one equal to the specified number of dimensions of the Latent Space (denoted as $\lambda$). The Fully Connected Network of the Decoder is identical to that of the Encoder, but in reverse order. Additionally, for the last Fully Connected Network of the Classifier Network, three fully connected layers were utilized, where the first two were assigned to 128 units and the final one equal to the number of classes of the corresponding dataset (Classification Layer). Finally, regarding the Activation Functions, for all the layers except for the output layers of the Encoder, Classifier and  the Decoder Networks, Rectified Linear Units are utilized, while for the aforementioned exceptions, Linear, Softmax and Sigmoid Activation Functions were employed, respectively.

    Each deep learning model was trained for 200 epochs and the model of highest validation accuracy during training was preserved. The mini-batch size and the optimizer are set to 128 and Adam~\cite{kingma2014adam}, with learning rate equal to $10^{-4}$, which is decreased by a factor of $1/3$ per 50 epochs, respectively. The remaining parameters for the classification algorithms, were kept to their default values, except for the number of neighbors parameter of the k-Nearest Neighbors Classifier, which was set to 3. 
    
    % I am not sure, this is what we should have done!!!!!!!
    
    % The proposed methodologies, were applied across all the datasets for different values of $\lambda$, and the most efficient performance is reported. The efficiency criterion, refers to the trade off between the value of $\lambda$ (dimensionality of the Latent Space of CSAE) and the performance. 
    % Specifically, small value of $\lambda$ and high performance is desired.

    The proposed methodologies, were applied across all the datasets for different values of $\lambda$ with minor performance variations confirming previous observations~\cite{MaiNgocHwang2020}.  We choose to report results for a relative small $\lambda$ value for which high classification accuracy can be obtained.
    The Keras implementation on the MNIST Dataset and detailed experimental results for different values of $\lambda$ of the proposed methodologies, along with additional visualizations can be found at the GitHub repository\footnote{see:\url{https://github.com/JohnNellas/CSAE}}. All the experiments were conducted on a server PC with Intel(R) Core(TM) i7-10700K CPU @ 3.80GHz, NVIDIA TITAN Xp 12 GB GPU, and 130GiB of RAM.

% OUR MODEL PARAMETERS MNIST AND FASHION MNIST 898,581 - > 0.9871 and 0.9117 accuracy
% VGG-5 (Spinal FC) - > 99 and 94 but with 3.630M parameters.
% our method achieves similar results but with 4.03x lower number of trainable parameters.

% VGG8B -> THE TEST ERROR IS REPORTED AND THUS -> ACCURACY = 100 - TEST ERROR*100 = 100 - 0.26 = 99.74 ~ 0.99
% VGG8B NUMBER OF PARAMETERS: 7.3M MNIST and Fashion MNIST
% our method achieves similar results but we achieve 8.12x lower number of trainable parameters

% COVID-19 PARAMETERS : 5,234,693
% BRAIM TUMOR: 5,329,545

\begin{table*}[!t]
		\caption{Performance Evaluation of CSAE and methodologies used for comparison on two benchmark image datasets and two biomedical image datasets.} \label{experimental_results_table}
		\begin{center}
			\ra{1.3}
			\resizebox{\linewidth}{!}{
				\begin{tabular}{@{}ccccccccc@{}}
					\toprule
					 & \multicolumn{2}{c}{MNIST} & \multicolumn{2}{c}{Fashion MNIST} & \multicolumn{2}{c}{Brain Tumor Dataset} & \multicolumn{2}{c}{SARS-COV-2 CT-Scans}\\
					
					 	& Accuracy	&  Weighted F1-Score	& Accuracy &	Weighted F1-Score &	Accuracy	& Weighted F1-Score & Accuracy	&  Weighted F1-Score \\
					 	 \midrule
				 	 LDA              & 0.8730 & 0.8726 & 0.8151 & 0.8159 & 0.9119 & 0.9126  & 0.8008 & 0.8007\\
				 	 CNN Classifier   & 0.9869 & 0.9869 & 0.9135 & 0.9132 & 0.9543 & 0.9540 & 0.9396 & 0.9396\\
				 	 AE + Class. Net. ($\lambda=\#classes$) & 0.6968 & 0.6759 & 0.5805 & 0.5362 & 0.4649 & 0.2951 & 0.4969  & 0.3520 \\
				 	 CSAE ($\lambda=2$) & 0.9751 & 0.9750 & 0.8959 & 0.8961 & \textit{\textbf{0.9575}} & \textit{\textbf{0.9574}} & 0.9436 & 0.9436\\
					 CSAE ($\lambda=\#classes$)             & \textit{\textbf{0.9871}} & \textit{\textbf{0.9871}} & \textit{\textbf{0.9117}} & \textit{\textbf{0.9114}} & 0.9510 & 0.9510 & \textit{\textbf{0.9436}} & \textit{\textbf{0.9436}}\\
					 VGG-5 (Spinal FC)~\cite{kabir2020spinalnet} & 0.9972 & - & 0.9468 & - & - & - & - & - \\
					 VGG8B~\cite{nokland2019training} & \textbf{0.9974} & - & \textbf{0.9547} & - & - & - & - & - \\
					 CapsNet~\cite{8451379}&   -       &       -        &        -       &        -       & 0.8656 &  - & -&-\\
					 CapsNet~\cite{8683759}&   -       &       -        &        -       &        -       & 0.9089 &  - & -&-\\
				% 	 VGG19~\cite{tazin2021robust}&   -       &       -        &        -       &        -       & 88.22 &  88.18  & -&-\\
				% 	 InceptionV3~\cite{tazin2021robust}&   -       &       -        &        -       &        -       & 91.00 &  90.98   & -&-\\
					 %MobileNetV2~\cite{tazin2021robust}&   -       &       -        &        -       &        -       & 92.00 &  92.00   & -&-\\
					 Contrastive Learning~\cite{wang2020contrastive} & - & - & - & - & - & - & 0.9083 & - \\
					 DenseNet201~\cite{Jaiswal2021Class} & - & - & - & - & - & - & \textbf{0.9574} & - \\
					 \bottomrule\\
					 \multicolumn{9}{l}{{\Large Best performance per dataset is highlighted using boldface text. Most efficient solution per dataset is denoted using}}\\
					 \multicolumn{9}{l}{{\Large boldface and italic text.}}\\
				\end{tabular}
			}
		\end{center}
\end{table*}

%   .

\subsection{Experimental Results}

% DOES THE PERFORMANCE ACHIEVED BY THE PROPOSED METHODOLOGY CAN BE CHARACTERIZED AS COMPETITIVE?

\begin{table*}[!b]
		\caption{Performance Comparison of the execution of traditional classification methodologies onto the flattened images and the Latent Representations constructed by CSAE .} \label{experimental_results_improving_table}
		\begin{center}
			\ra{1.3}
			\resizebox{\linewidth}{!}{
				\begin{tabular}{@{}ccccccccc@{}}
					\toprule
					 & \multicolumn{2}{c}{MNIST} & \multicolumn{2}{c}{Fashion MNIST} & \multicolumn{2}{c}{Brain Tumor Dataset} & \multicolumn{2}{c}{SARS-COV-2 CT-Scans}\\
					
					 	& Accuracy	&  Weighted F1-Score	& Accuracy &	Weighted F1-Score &	Accuracy	& Weighted F1-Score & Accuracy	&  Weighted F1-Score \\
					 	 \midrule
				 	 k-NN             & 0.9580 & 0.9579 & 0.8915 & 0.8911 & 0.8874 & 0.8836 & 0.8651 & 0.8647\\
				 	 CSAE L.S. + kNN ($\lambda=2$) & 0.9785 & 0.9784 & 0.9245 & 0.9241 & 0.9624 & 0.9625 & 0.9657 & 0.9657\\
				 	 CSAE L.S. + kNN ($\lambda=\#classes$)  & \textit{\textbf{0.9926}} & \textit{\textbf{0.9925}} & \textit{\textbf{0.9309}} & \textit{\textbf{0.9303}} & \textit{\textbf{0.9657}} & \textit{\textbf{0.9656}} & \textit{\textbf{0.9657}} & \textit{\textbf{0.9657}}\\
				 	 \midrule
				 	 SVM.             & 0.9660 & 0.9660 & 0.8836 & 0.8828 & 0.9135 & 0.9120 & 0.9336 & 0.9335\\
				 	 CSAE L.S. + SVM ($\lambda=2$) & 0.9758 & 0.9758 & 0.8959 & 0.8964 & \textbf{0.9608} & \textbf{0.9607} & 0.9436 & 0.9436\\
					 CSAE L.S. + SVM ($\lambda=\#classes$)  & \textbf{0.9872} & \textbf{0.9871} & \textbf{0.9157} & \textbf{0.9152} & 0.9510 & 0.9510 & \textbf{0.9436} & \textbf{0.9436}\\
					 \midrule
					 GNB              & 0.5240   & 0.4772   & 0.5706   & 0.5398   & 0.7406   & 0.7329   & 0.7364 & 0.7285\\
					 CSAE L.S. + GNB ($\lambda=2$) & 0.9161 & 0.9167 & 0.8244 & 0.8264 & 0.9200 & 0.9196 & 0.9456 & 0.9456\\
					 CSAE L.S. + GNB ($\lambda=\#classes$)  & \textbf{0.9742} & \textbf{0.9742} & \textbf{0.8743} & \textbf{0.8724} & \textbf{0.9396} & \textbf{0.9396} & \textbf{0.9456} & \textbf{0.9456}\\
					 \bottomrule\\
					 \multicolumn{9}{l}{{\Large Best performance per dataset and classification method is highlighted using boldface text. Best performance per dataset }}\\
					 \multicolumn{9}{l}{{\Large across classification methods is denoted using  boldface and italic text. }}
				\end{tabular}
			}
		\end{center}
\end{table*}

% 

% \begin{figure}[b!]
%   \centering
%   \includegraphics[width=\textwidth]{./images/covid_19_network_embedded_space_decision_boundary_2D_latent_dim_2_classes.png}
%   \caption{The Latent Space of CSAE for the images in the test set of the SARS-COV-2 CT-Scans Dataset and the decision boundary of the Classifier drawn on the corresponding Latent Space. A bold coloured point corresponds to the ground truth class of the corresponding latent representation, while a point with lower opacity corresponds to the class predicted by the network.} \label{fig:decision_boundary}
% \end{figure}

    % Simeiosi apo Gianni : Isos aksizei na tonisoume oti auta ta apotelesmata simeiothikan
    % xwris prosthiki kapoiou epipleon regularization gia na tonisoume to entono regularization
    % pou kanei to unsupervised loss!!!
    The experimental results, regarding the first set of comparisons, are reported in Table~\ref{experimental_results_table}. We observe that CSAE performance surpasses the methods used for comparisons while also achieving competitive results against well established methods from the literature.
    In more detail, for the Fashion MNIST datasets, even though CSAE achieved inferior performance than the methods presented in ~\cite{kabir2020spinalnet, nokland2019training}, it constitutes a significantly smaller model with 4.03 and 8.12 times less parameters respectively.
    Regarding the Braim Tumor Dataset, CSAE outperformed any other method, including those presented in~\cite{8451379, 8683759}. In addition, for the SARS-COV-2 CT-Scans dataset, CSAE achieved only slightly worse performance than the methodology presented in \cite{Jaiswal2021Class}, but still, the proposed methodology reached this performance with 3.85 times less parameters, allowing us to dismiss any need for transfer learning. 
    % in contrast with the study used for comparison.
    Most importantly, we be observed that the proposed optimization strategy of the reconstruction and classification error leads to an improvement over the scheme where two procedures are performed independently or by only using a single CNN classifier.
    % Finally, we interestingly observe that the Linear Discriminant Analysis algorithm, which is a methodology for supervised dimensionality reduction and classification, achieves descent results even on biomedical image datasets, which are real data and thus characterized by higher complexity.

\begin{figure}[b!]
          \centering
          \subfloat[MNIST]{\includegraphics[width=0.33\linewidth]{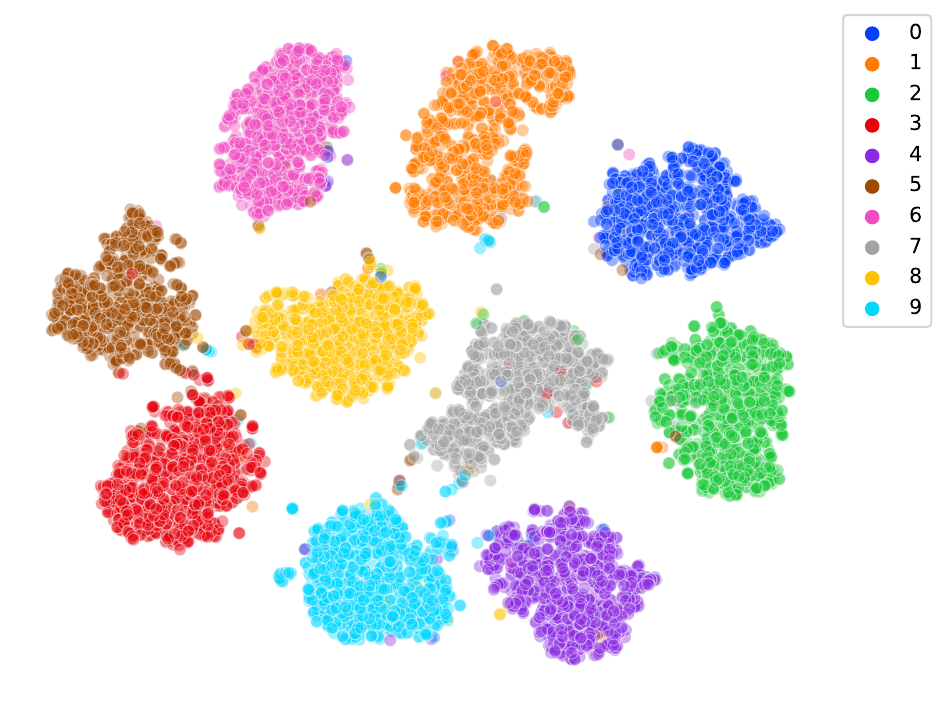} \label{fig:mnist_latent_space}}
          \subfloat[Fashion MNIST]{\includegraphics[width=0.33\linewidth]{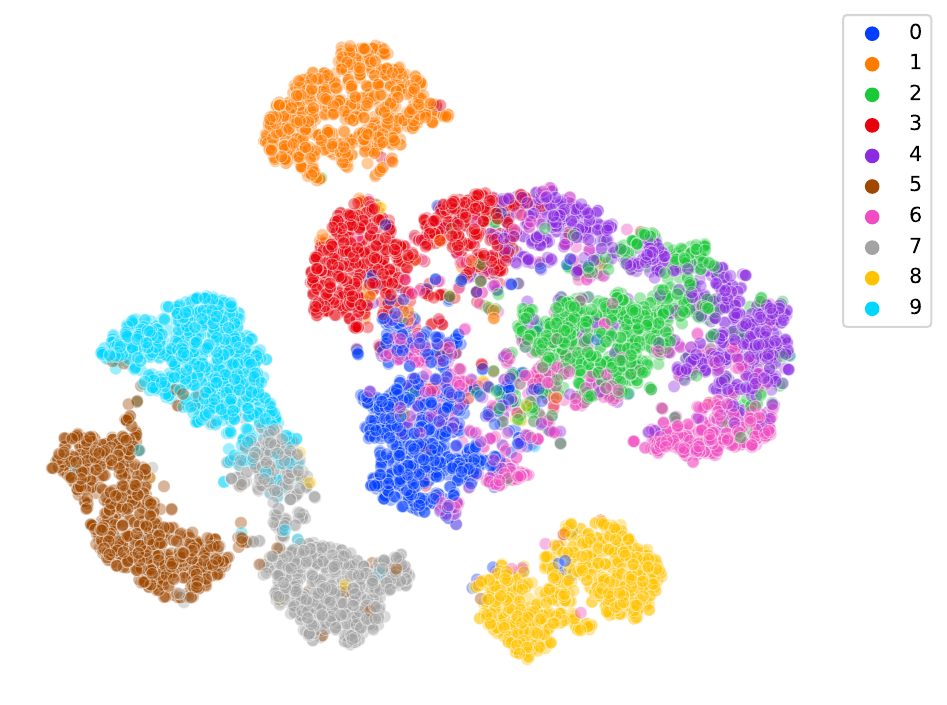} \label{fig:fmnist_latent_space}}
          \subfloat[Brain Tumor]{\includegraphics[width=0.33\linewidth]{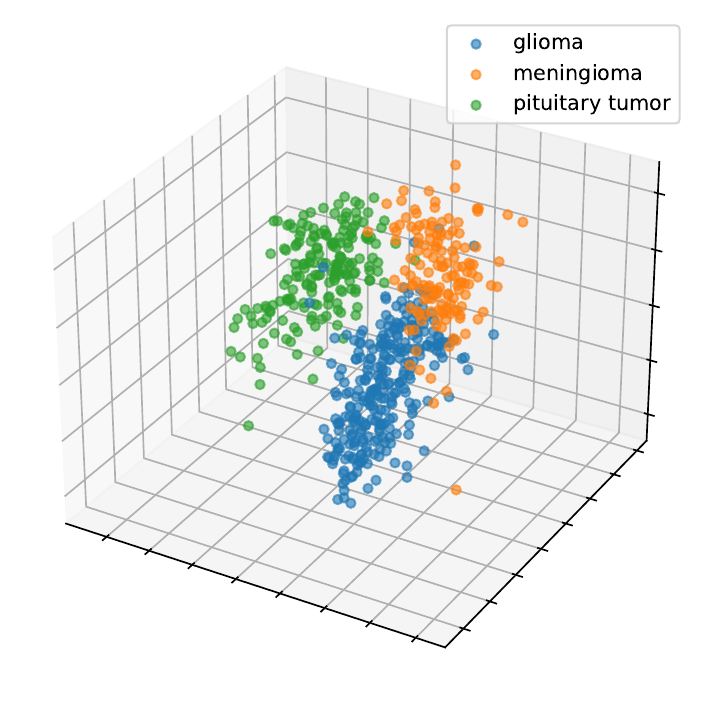} \label{fig:brain_latent_space}}
          \caption{A visualization of the Latent Space created by CSAE, of the images contained in the test set of MNIST (a), Fashion MNIST (b) and Brain Tumor (c) Datasets, for $\lambda=\#classes$. The visualization of the Latent Space for the first two datasets was created using the t-SNE algorithm, while for the latter the original latent space is plotted. Finally, different colours correspond to different ground truth classes.} \label{fig:latent_space}
    \end{figure}

    The second set of experimental results are reported in Table~\ref{experimental_results_improving_table}. It can be observed that the performance of the traditional classification methodologies was significantly improved, when they were applied onto the optimized for classification latent representations produced by CSAE.
    % Additionally, notice, the significant performance improvement of the naive bayes classifier, when applied onto these representations in comparison with the original flattened images.
    The best performance across methods and metrics, was achieved by the k-Nearest Neighbors classifier. Interestingly, it even surpasses that of CSAE in Table~\ref{experimental_results_table}.
    %
    % T-SNE citation added!
    The aforementioned results can be visually justified by investigating two dimensional representations of the Latent Spaces constructed by CSAE, retrieved by the t-SNE algorithm \cite{van2008visualizing}.
    % , for the images contained in the test sets of the utilized datasets, which are presented in the
    As shown in Figure~\ref{fig:latent_space}
    % and right Subfigure of the Figure~\ref{fig:covid19_CAEVSCSAE}.
    % !!!!!!!!!!!!!!!!!!!!!!!!!!! REFER TO IT LATER
   , we observe that points of the same class form dense neighborhoods of points.
%   and thus a k-Nearest Neighbors classifier can perform adequately in this case.
    
    % FMNIST
    
    % A paragraph about the explainability, the decision boundary offers explainability, "A sneak peak into the black box". See all the data in one place rather than individually, with overlay.

    \begin{figure}
    	\centering
    	\includegraphics[width=\textwidth]{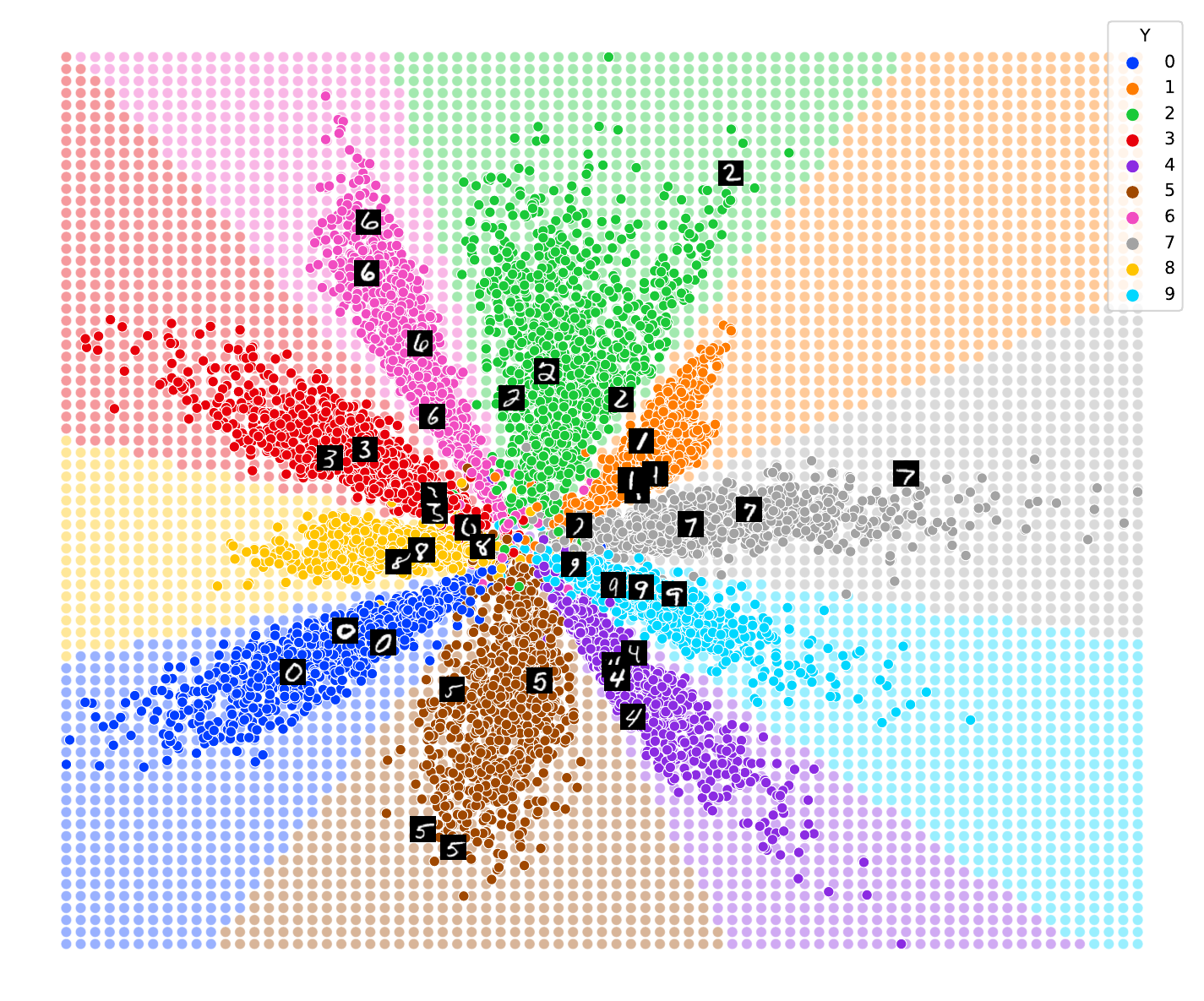}
    	\caption{The Latent Space of CSAE for the images in the test set of the MNIST Dataset and the decision boundary of the Classifier drawn on the corresponding Latent Space. A bold coloured point corresponds to the ground truth class of the corresponding latent representation, while a point with lower opacity corresponds to the class predicted by the network. Four random embeddings per class were replaced by the original image.}\label{fig:scatter_images_datasets_MNIST}
    \end{figure}
    \begin{figure}
    	\centering
    	\includegraphics[width=\textwidth]{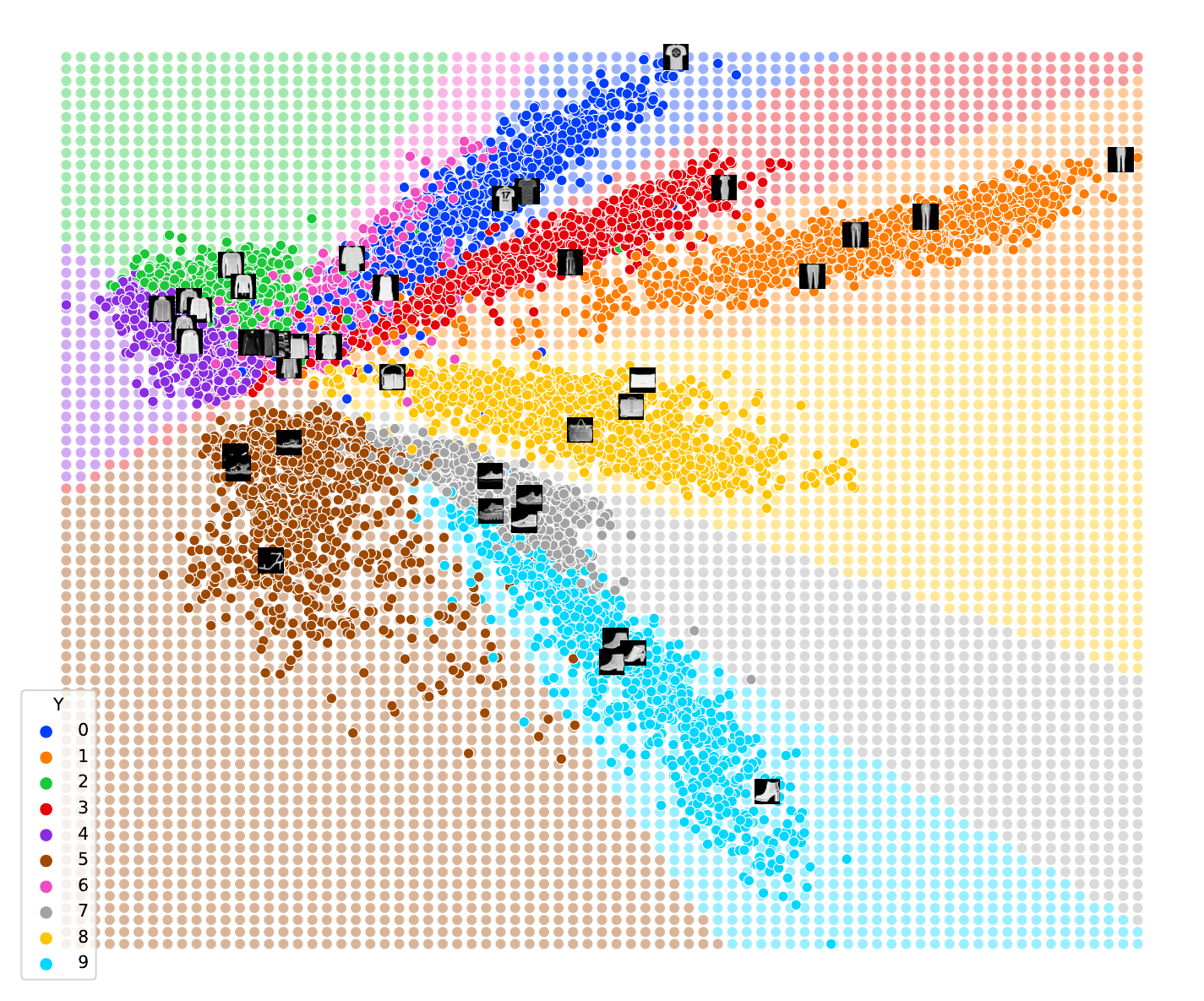}
    	\caption{The Latent Space of CSAE for the images in the test set of the Fashion MNIST Dataset and the decision boundary of the Classifier drawn on the corresponding Latent Space. A bold coloured point corresponds to the ground truth class of the corresponding latent representation, while a point with lower opacity corresponds to the class predicted by the network. Four random embeddings per class were replaced by the original image}\label{fig:scatter_images_datasets_FMNIST}
    \end{figure}
    \begin{figure}
    	\centering
    	\includegraphics[width=\textwidth]{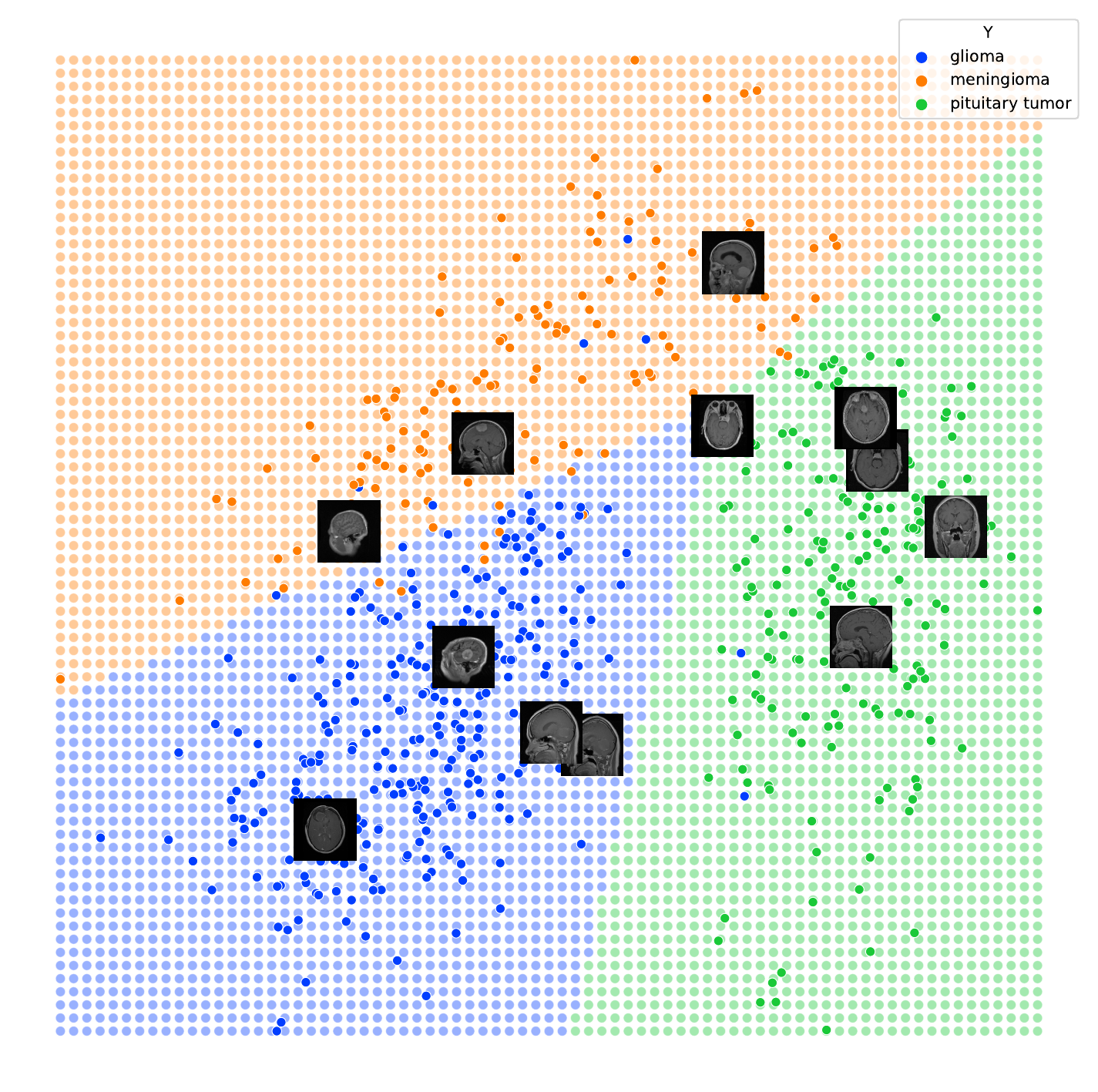}
    	\caption{The Latent Space of CSAE for the images in the test set of the Brain Tumor Dataset and the decision boundary of the Classifier drawn on the corresponding Latent Space. A bold coloured point corresponds to the ground truth class of the corresponding latent representation, while a point with lower opacity corresponds to the class predicted by the network. Four random embeddings per class were replaced by the original image}\label{fig:scatter_images_datasets_CANCER}
    \end{figure}
    \begin{figure}
    	\centering
    	\includegraphics[width=\textwidth]{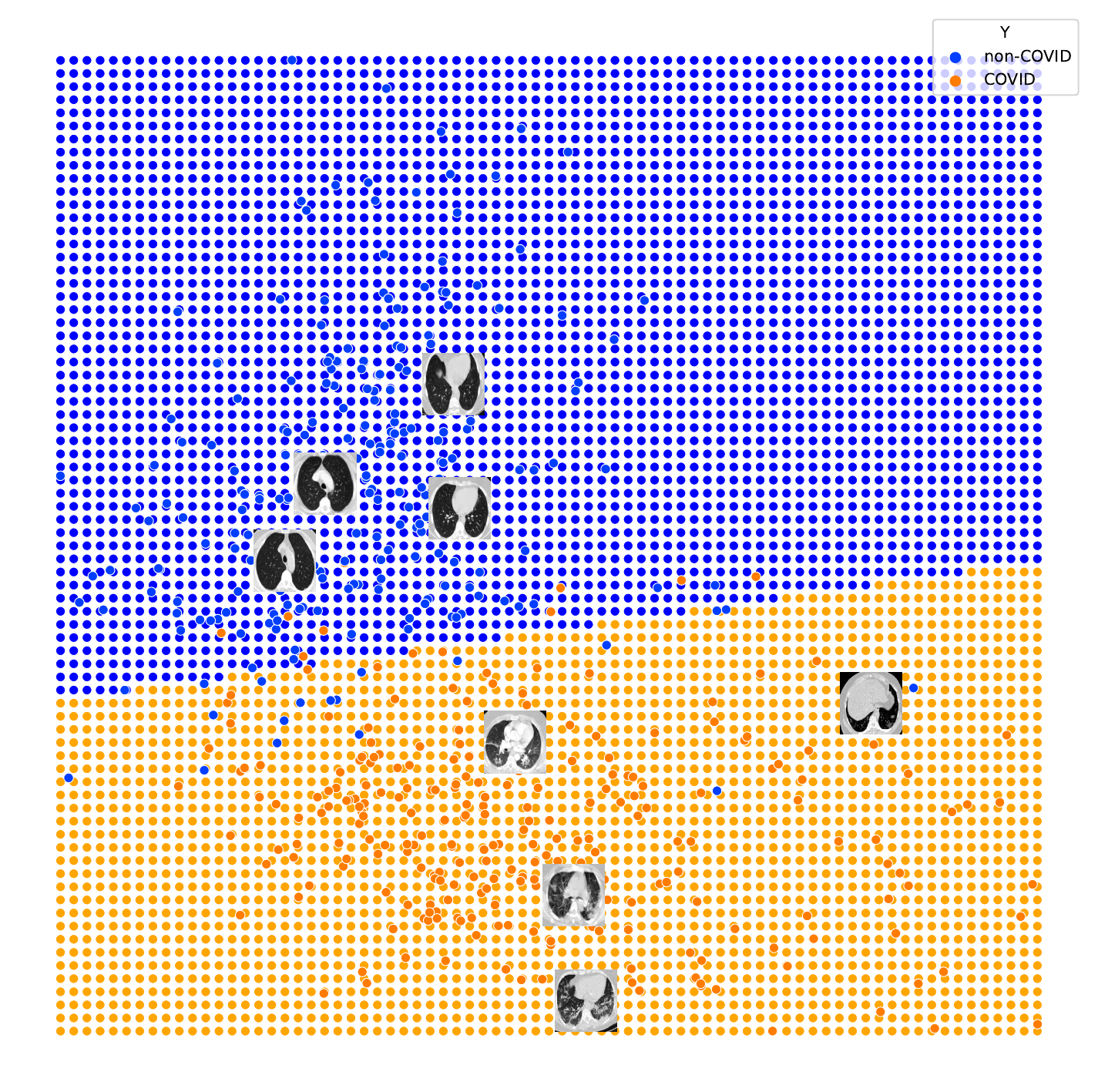}
    	\caption{
    	The Latent Space of CSAE for the images in the test set of the SARS-COV-2 CT-Scans Dataset and the decision boundary of the Classifier drawn on the corresponding Latent Space. A bold coloured point corresponds to the ground truth class of the corresponding latent representation, while a point with lower opacity corresponds to the class predicted by the network. Four random embeddings per class were replaced by the original image}\label{fig:scatter_images_datasets_COVID_19}
    \end{figure}

    % A FIGURE OF THE FASHION MNIST DECISION BOUNDARY + LATENT SPACE + SOME ORIGINAL IMAGES
    % \begin{figure}[b!]
    %     \centering
    %     \subfloat[MNIST]{\includegraphics[width=0.5\linewidth]{./images/mnist_network_embedded_space_decision_boundary_2D_latent_dim_10_classes.png}}
    %     \subfloat[Fashion MNIST]{\includegraphics[width=0.5\linewidth]{./images/fmnist_network_embedded_space_decision_boundary_2D_latent_dim_10_classes.png}}\\
    %     \subfloat[Brain Tumor Dataset]{\includegraphics[width=0.5\linewidth]{./images/cancer_dataset_gray_network_embedded_space_decision_boundary_2D_latent_dim_3_classes.png}}
    %     \subfloat[SARS-COV-2 CT-Scans]{\includegraphics[width=0.5\linewidth]{./images/covid_19_network_embedded_space_decision_boundary_2D_latent_dim_2_classes.png}}
    %     \caption{The Scatter Plot of the Latent Space of CSAE for the images in the test set of the datasets used for evaluation, along with the corresponding classification boundary. Randomly selected 100 embeddings were replaced by the corresponding original image.}
    %     \label{fig:scatter_images_datasets}
    % \end{figure}

    \subsubsection{Visualization}
    
    The proposed Convolutional Supervised Autoencoder, allow both the visualization of the generated Latent Space and the decision boundary of the Classifier Network. Specifically, we set $\lambda=2$ to retrieve the two dimensional Latent Space that is subsequently provided as input to the fully connected network of the classifier. We visualize decision regions and boundaries by generating a coloured scatter grid of points where each colour corresponds to a prediction class.
    % by creating a grid of points in the two dimensional Latent Space ($\lambda=2$), providing them as input to the fully connected network of the classifier, and then create a coloured scatter plot of the points in the grid, where the colour will correspond to the class prediction of each grid point, will provide a visualization of the decision boundary of the classifier onto the latent space.
    % 
    A scatter plot of the Latent Representations of the images contained in the test set of the MNIST, Fashion MNIST, Brain Tumor and SARS-COV-2 CT-Scan datasets, and the decision boundary of the Classifier drawn on the corresponding Latent Space, is presented in Figures~\ref{fig:scatter_images_datasets_MNIST},~\ref{fig:scatter_images_datasets_FMNIST},~\ref{fig:scatter_images_datasets_CANCER} and~\ref{fig:scatter_images_datasets_COVID_19}, correspondingly.
    A bold coloured point corresponds to the ground truth class of the corresponding latent representation, while a point with lower opacity corresponds to the class predicted by the network.
    We observe that, decision regions in the Latent Space created by CSAE, are almost linearly separable, and the classifier converged to a linear decision boundary.
    This observation confirms the objective of most data transformation methodologies, where a non linear data transformation creates a space where the data are linearly separable. Most importantly, in this case the visualization offers the much requested explainability, since it allows the realization of the decision that the network makes in order to classify the input points.
    % This observation strengthens the indication that a non linear data transformation creates a space where the data are linearly separable. Finally, the decision boundary visualization offers explainability about the classification behavior of the network, because allows the realization of the decision that the network makes in order to classify the input points.

    % ADDED References
    % cheng2016retrieval: (Brain Tumor) same class large variability, different classes can have similar appearances!
    % hani2020covid: (COVID-19): difficult to distinguish between COVID-19 pneumonia and other viral pneumonias due to CT features being overlapped.
    % JIA2021104425: (COVID-19)
        % 1. changes occur in CXR and CT images before the beginning of COVID-19 
        % 2. symptoms of COVID-19 and other lung diseases can be similar in their very early stage
    
    An example of the explainability provided by the aforementioned scatter plots can be illustrated by randomly replacing data points with the corresponding original images. Then we can simultaneously visually examine their pairwise distances and their distance from the decision boundary with respect to their visual characteristics. Apparently, images that lookalike tend to be closer to each other confirming the data structure preservation capability of the proposed methodology. Visual investigation of the images found across the decision boundary can be extremely beneficial for real world bio-medicine applications where class membership is often not easily distinguishable \cite{cheng2016retrieval,hani2020covid,JIA2021104425}.

\begin{figure}[t!]
      \centering
      \subfloat[MNIST]{\includegraphics[width=0.5\linewidth]{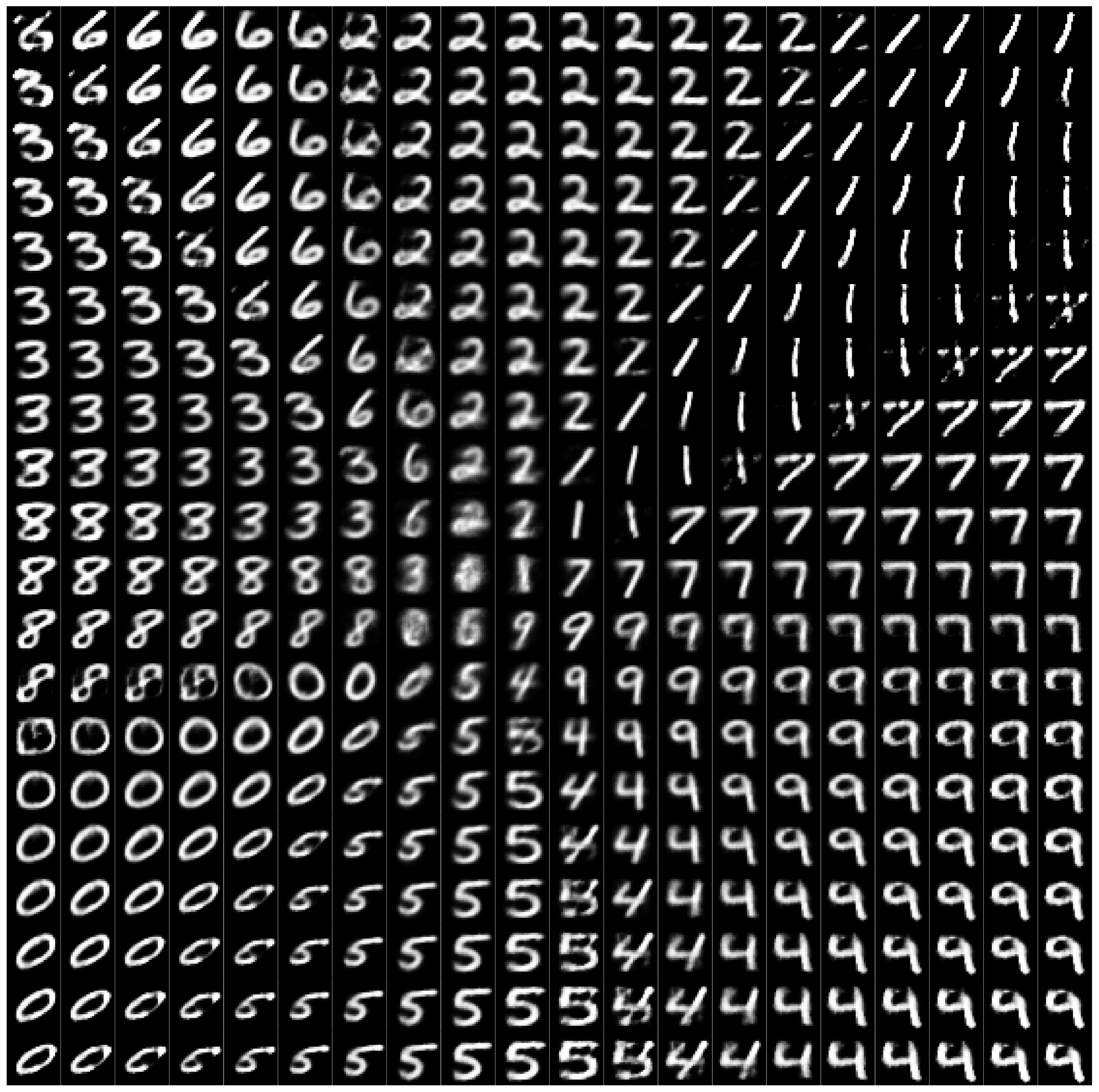} \label{fig:mnistdecodedpoints}}
      \subfloat[Fashion MNIST]{\includegraphics[width=0.5\linewidth]{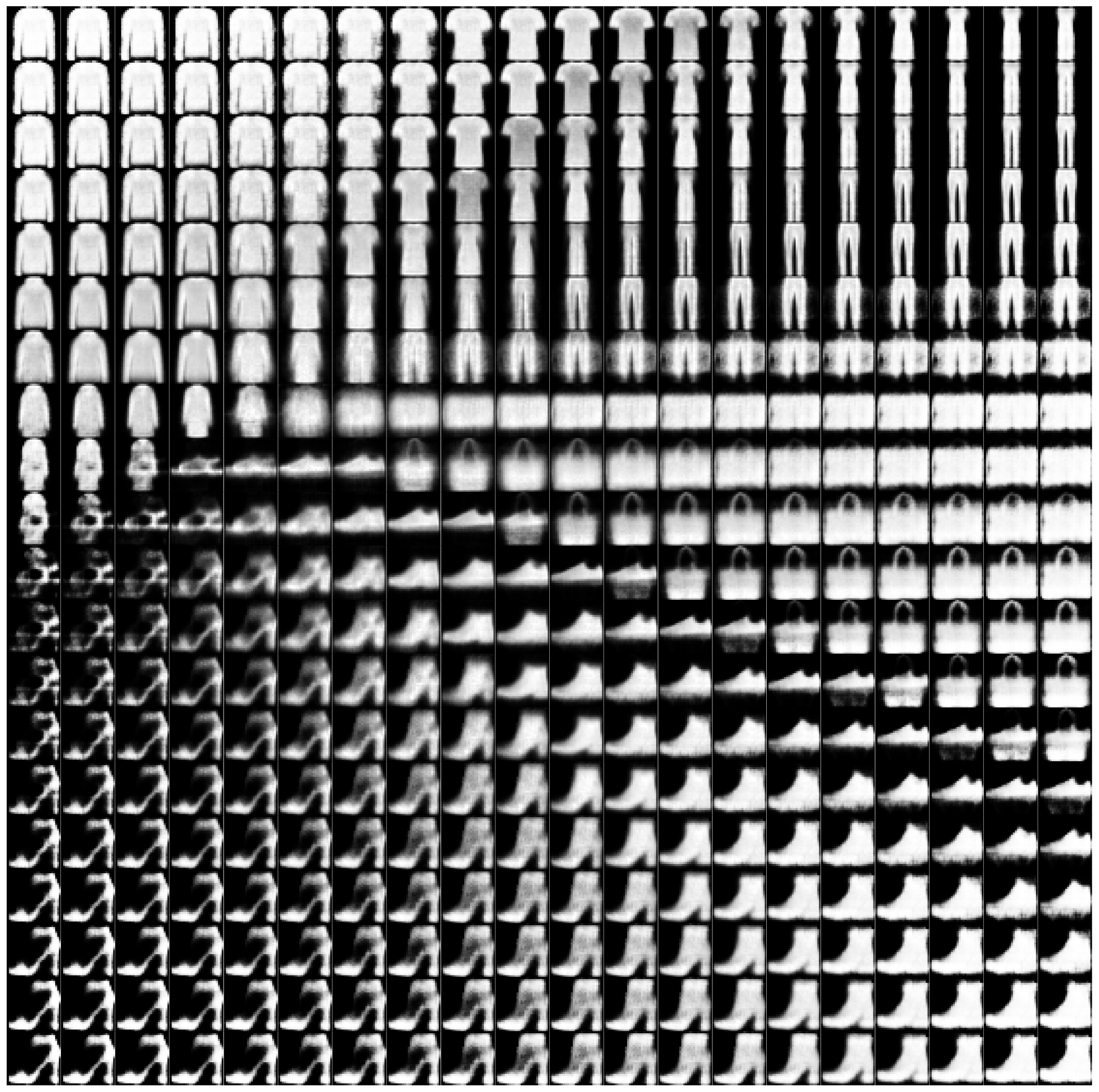} \label{fig:fmnistdecodedpoints}}
      \caption{A grid of images from the MNIST (left) and Fashion MNIST (right) datasets, decoded from a grid of points in the Latent Space of CSAE} \label{fig:decoded_points}
\end{figure}

    Finally, we further examine the information that the network has captured regarding the data in the Latent Space of CSAE by providing a grid of points from the Latent Space as input to the Decoder Network. A decoded grid of points from the two dimensional ($\lambda=2$) Latent Space of CSAE, for the MNIST and Fashion MNIST datasets, is presented in Figure~\ref{fig:decoded_points}. We observe that, except from the label information, the network has also captured the rotation and intensity information of the MNIST and Fashion MNIST dataset respectively.
    
    % A final study that can be performed, includes the usefulness of information that the network has captured regarding the data in the Latent Space of CSAE. The latter can be performed by providing a grid of points from the Latent Space as input to the Decoder Network and examine the resulting images. A decoded grid of points from the two dimensional ($\lambda=2$) Latent Space of CSAE, for the MNIST and Fashion MNIST datasets, is presented in Figure~\ref{fig:decoded_points}. From the latter it is observed that, except from the label information, the network has also captured the rotation and intensity information of the MNIST and Fashion MNIST dataset respectively.

% \begin{figure}[t!]
%   \centering
%   \subfloat[MNIST]{\includegraphics[width=0.5\linewidth]{images/mnist_decision_boundary_grid_of_points.png} \label{fig:mnist_decision_boundary}}
%   \subfloat[Fahsion MNIST]{\includegraphics[width=0.5\linewidth]{images/fmnist_decision_boundary_grid_of_points.png} \label{fig:fmnist_decision_boundary}}\\
%   \subfloat[Brain Tumor]{\includegraphics[width=0.5\linewidth]{images/cancer_dataset_decision_boundary_grid_of_points.png} \label{fig:cancer_decision_boundary}}
%   \subfloat[COVID-19 CT Scans]{\includegraphics[width=0.5\linewidth]{images/covid_19_decision_boundary_grid_of_points.png} \label{fig:covid_19_decision_boundary}}
%   \caption{The Decision Boundary drawn on the Latent Space of CSAE for the binary MNIST (a), binary Fashion MNIST (b) binary Brain Tumor Dataset (c) and the COVID-19 CT Scans (d) Datasets.} \label{fig:decision_boundary}
% \end{figure}

 % ============until here=====================
\section{Conclusions}

    In this study, a novel supervised dimensionality reduction and classification methodology is proposed, which is constituted by a Convolutional Autoencoder for dimensionality reduction and a classifier for the classification of the latent representations. Its main characteristic is that it concurrently optimizes not only the reconstruction but also the classification error. This method is entitled Convolutional Supervised Autoencoder (CSAE). In addition, we consider the produced Latent Space of the proposed methodology as optimized for classification, and thus we argue that its latent representations can be provided as inputs to a trainable classifier to significantly improve their performance. To support the aforementioned claims, a thorough study regarding the Latent Space and the classification behaviour of CSAE is performed. The experimental results on two benchmark and two biomedical image datasets, showed that CSAE, achieved competitive classification performance against state of the art, while surpassing alternative methodological scenarios. Simultaneously, it offers a much more efficient solution in terms of parameters count. It is also observed that the performance of traditional classification algorithms was indeed improved when they were applied onto the latent representations of CSAE. Most importantly, motivated by the visualization capabilities of CSAE we investigated the explainability perspective, which adds greater value to the proposed methodology. We specifically observed that the resulting decision boundaries of the classifier converged to a linear hyperplanes. To that end, we highlight our interest in further investigating similar architectures that enable us to visualized and provide wider explainability in time series classification tasks.
    
    % Regarding the data structure preservation in the latent space, it was observed that, similar images correspond to closer embeddings than dissimilar images, and thus concluding that data structure is preserved in the Latent Space of CSAE. Furthermore, it was observed that the consideration of the label information onto the construction of the Latent Space, greatly increased the separability of the data. Finally, the Latent Space of CSAE was able to  capture some useful information regarding the input data, such as the rotation information.

\section*{Acknowledgements}
This project has received funding from the Hellenic Foundation for Research and Innovation (HFRI), under grant agreement No 1901.

\printbibliography
\end{document}